%% file: main.tex
\documentclass[10pt,twocolumn,letterpaper]{article}

\usepackage{cvpr}
\usepackage{times}
\usepackage{epsfig}
\usepackage{graphicx}
\usepackage{amsmath}
\usepackage{amssymb}
\usepackage{subfig}
\usepackage{xcolor}
\usepackage{soul}
\usepackage{mathtools}
\usepackage{multirow}
\usepackage{hhline}

\input{symbols.tex}

\usepackage[pagebackref=true,breaklinks=true,letterpaper=true,colorlinks,bookmarks=false]{hyperref}

\cvprfinalcopy %

\def\cvprPaperID{10262} %

\newcommand{\figref}[1]{Figure \ref{figure:#1}}
\newcommand{\secref}[1]{\S \ref{section:#1}}
\renewcommand{\eqref}[1]{(\ref{equation:#1})}
\newcommand{\tabref}[1]{Table \ref{table:#1}}

\newcommand{\self}{\mathtt{self}}
\newcommand{\other}{\mathtt{other}}

\newcommand{\free}{\mathtt{free}}

\newcommand{\obj}{o}
\newcommand{\objidx}{m}
\newcommand{\nontargetobj}{{\widetilde{\obj}}}
\newcommand{\nontargetobjidx}{n}

\newcommand{\point}{p}
\newcommand{\pointidx}{q}
\newcommand{\voxel}{v}
\newcommand{\voxelsize}{s}
\newcommand{\voxelorig}{l}
\newcommand{\voxelidx}{k}
\newcommand{\pointtovoxel}{u}
\newcommand{\occupancy}{o}
\newcommand{\impen}{{\mathtt{impen}}}

\begin{document}

\title{MoreFusion: Multi-object Reasoning for 6D Pose Estimation \\ from Volumetric Fusion}

\author{Kentaro Wada, Edgar Sucar, Stephen James, Daniel Lenton, Andrew J. Davison\\
Dyson Robotics Laboratory, Imperial College London\\
{\tt\small \{k.wada18, e.sucar18, slj12, djl11, a.davison\}@imperial.ac.uk}}

\maketitle

\begin{abstract}
\input{abstract/abstract.tex}
\end{abstract}
\vspace{-1em}

\input{introduction/introduction.tex}

\input{related_work/related_work.tex}

\input{methods/methods.tex}

\input{experiments/experiments.tex}

\input{conclusions/conclusions.tex}

\vspace{-2mm}
\input{acknowledgements/acknowledgements.tex}

{\small
\bibliographystyle{ieee_fullname}
\bibliography{robotvision}
}

\end{document}

%% file: symbols.tex
\newcommand{\ignore}[1]{}

\newcommand{\ba}{\begin{array}}
\newcommand{\ea}{\end{array}}
\newcommand{\bc}{\begin{center}}
\newcommand{\ec}{\end{center}}
\newcommand{\be}{\begin{enumerate}}
\newcommand{\ee}{\end{enumerate}}
\newcommand{\bea}{\begin{eqnarray}}
\newcommand{\eea}{\end{eqnarray}}
\newcommand{\beas}{\begin{eqnarray*}}
\newcommand{\eeas}{\end{eqnarray*}}
\newcommand{\beq}{\begin{equation}}
\newcommand{\eeq}{\end{equation}}
\newcommand{\bfig}{\begin{figure}}
\newcommand{\efig}{\end{figure}}
\newcommand{\bi}{\begin{itemize}}
\newcommand{\ei}{\end{itemize}}
\newcommand{\bpic}{\begin{picture}}
\newcommand{\epic}{\end{picture}}
\newcommand{\btabular}{\begin{tabular}}
\newcommand{\etabular}{\end{tabular}}
\newcommand{\btable}{\begin{table}}
\newcommand{\etable}{\end{table}}

\newcommand{\es}{\vfill
                 \rule[-6mm]{170mm}{0.7mm} \\
                 \redw{{\tiny
		  \hfill S-\theslide}}
                 \end{slide}}

\newcommand{\matxx}[1]{{\mathtt #1}}
\newcommand{\vecXX}[1]{{\mathbf {#1}}}

\def \hbar {{\bar{h}}}

\def \vect {{\vecXX{t}}}

\def \vecx {{\vecXX{x}}}

\def \vecthat {\hat{\vecXX{t}}}

\def \matR {{\matxx{R}}}

\def \matRhat {\hat{\matxx{R}}}

\makeatletter
\renewcommand*\env@matrix[1][*\c@MaxMatrixCols c]{%
  \hskip -\arraycolsep
  \let\@ifnextchar\new@ifnextchar
  \array{#1}}
\makeatother

%% file: abstract/abstract.tex
Robots and other smart devices need efficient object-based scene
representations from their on-board vision systems to reason about
contact, physics and occlusion. Recognized precise object models will
play an important role alongside non-parametric reconstructions of
unrecognized structures. We present a system which can estimate the
accurate poses of multiple known objects in contact and occlusion from
real-time, embodied multi-view vision. Our approach makes 3D object
pose proposals from single RGB-D views, accumulates pose
estimates and non-parametric occupancy information from multiple views
as the camera moves, and performs joint optimization to estimate
consistent, non-intersecting poses for multiple objects in contact.

We verify the accuracy and robustness of our approach experimentally
on 2 object datasets: YCB-Video, and our own challenging Cluttered YCB-Video. We demonstrate a real-time robotics application where a robot arm precisely and orderly disassembles
complicated piles of objects, using only on-board RGB-D vision.

%% file: introduction/introduction.tex
\section{Introduction}

\begin{figure}[t]
  \centering
  \subfloat[Pose Estimation]{
    \includegraphics[width=0.90\linewidth]{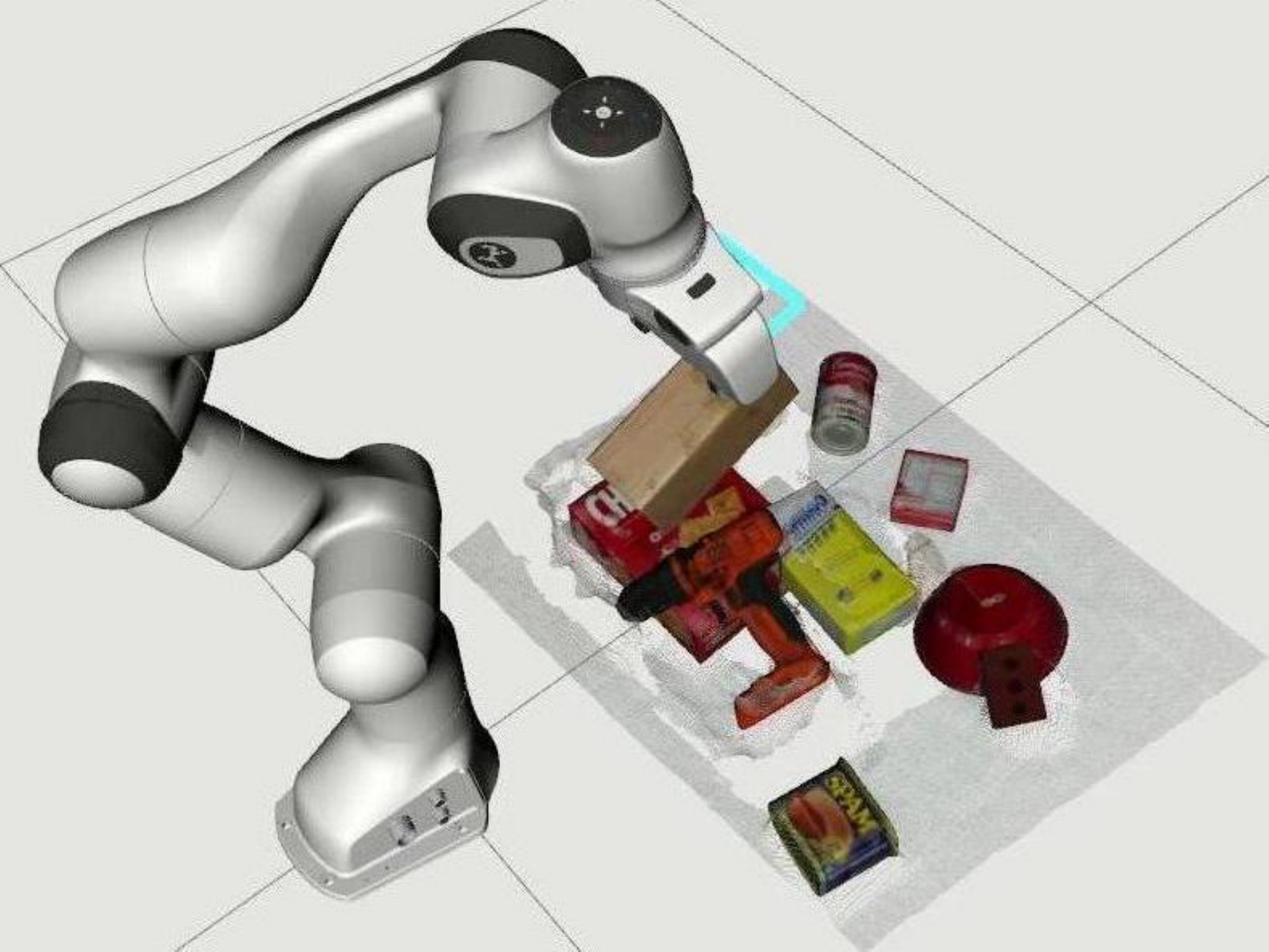}
    \label{figure:introduction:rviz_pose}
  }\\
  \subfloat[Volumetric Fusion]{
    \includegraphics[width=0.43\linewidth]{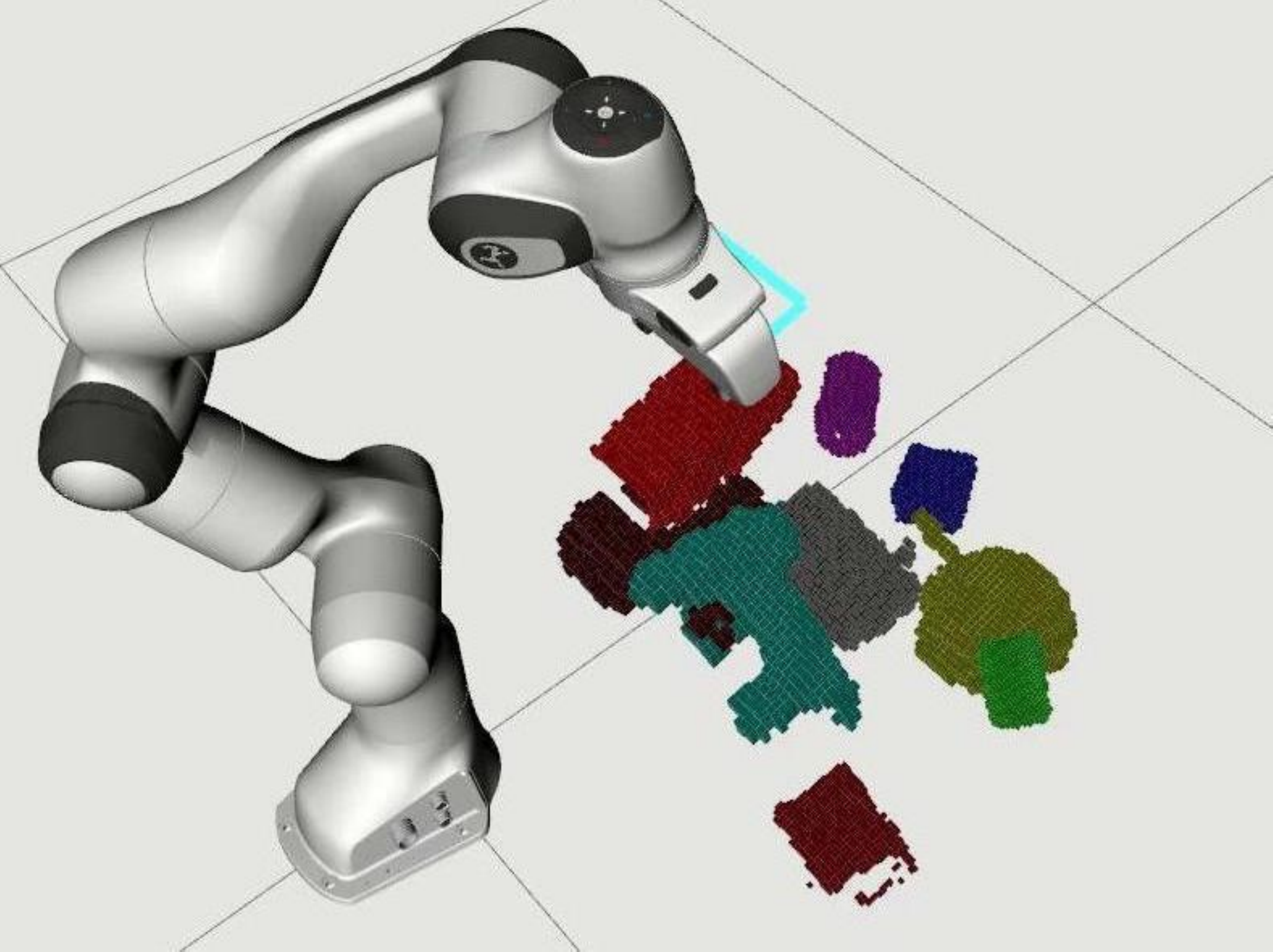}
    \label{figure:introduction:rviz_volumetric}
  }
  \subfloat[Real Scene]{
    \includegraphics[width=0.46\linewidth]{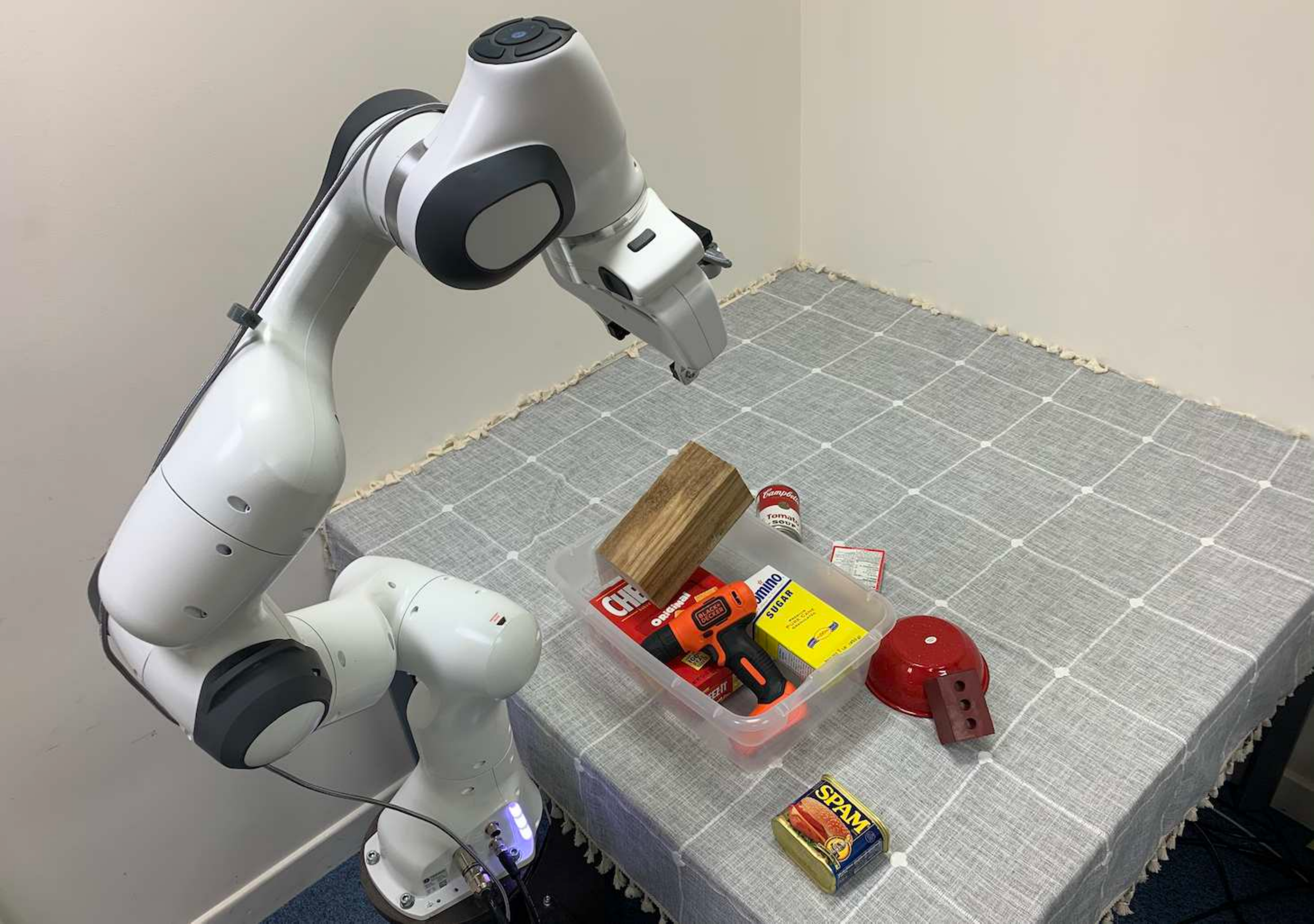}
  }
  \caption{
  \textbf{MoreFusion} produces accurate 6D object pose predictions by explicitly reasoning about occupied and free space via a volumetric map. We demonstrate the system in a real-time robot grasping application.
  }
  \label{figure:introduction}
  \vspace{2mm} \hrule \vspace{-4mm}
\end{figure}

Robots and other smart devices that aim to perform complex tasks such as
precise manipulation in cluttered environments need to capture information from
their cameras that enables reasoning about contact, physics and occlusion among
objects.  While it has been shown that some short-term tasks can be
accomplished using end-to-end learned models that connect sensing to action, we
believe that extended and multi-stage tasks can greatly benefit from persistent
3D scene representations.

\begin{figure*}[t]
  \centering
  \includegraphics[width=\linewidth]{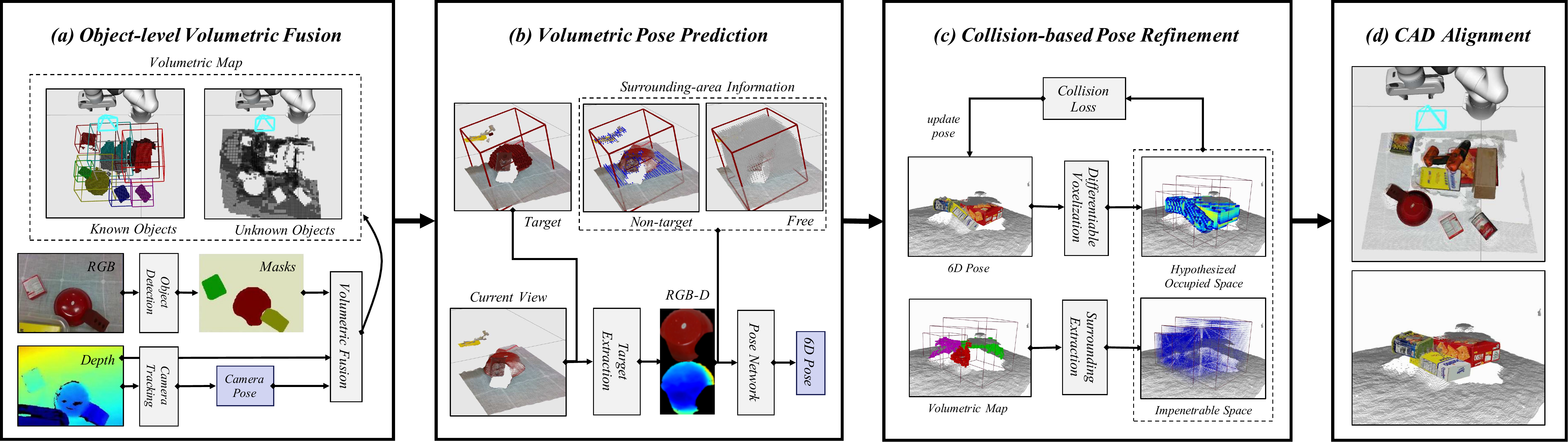}

  \caption{{\bf Our 6D pose estimation system.} Object segmentation masks from
    RGB images are fused into a volumetric map , which denotes both occupied
    and free space (a). This volumetric map is used along with RGB-D data of
    a target object crop to make an initial 6D pose prediction (b). This pose is
    then refined via differentiable collision checking (c) and then used as part of a
    CAD alignment stage to enriches the volumetric map (d).}

  \label{figure:system}
  \vspace{2mm} \hrule \vspace{-4mm}
\end{figure*}

Even when the object elements of a scene have known models, inferring the
configurations of many objects that are mutually occluding and in contact is
challenging even with state-of-the-art detectors. In this paper we present a
vision system that can tackle this problem, producing a persistent 3D
multi-object representation in real-time from the multi-view images of a single
moving RGB-D camera. Our system has four main components, as highlighted in
Figure~\ref{figure:system}: 1) 2D object detection is fed to object-level
fusion to make volumetric occupancy map of objects.  2) A pose prediction
network that uses RGB-D data and the surrounding occupancy grid makes 3D object
pose estimates.  3) Collision-based pose refinement jointly optimizes the poses
of multiple objects with differentiable collision checking.  4) The
intermediate volumetric representation of objects are replaced with
information-rich CAD models.

Our system takes full advantage of depth information and multiple views to
estimate mutually consistent object poses. The initial rough volumetric
reconstruction is upgraded to precise CAD model fitting when models can be
confidently aligned without intersecting with other objects.  This visual
capability to infer the poses of multiple objects with occlusion and contact
enables robotic planning for pick-and-place in a cluttered scene \eg removing
obstacle objects for picking the target red box in \figref{introduction}.

In summary, the main contributions of this paper are:
\begin{itemize}
  \setlength\itemsep{0em}
  \item {\bf Pose prediction with surrounding spatial awareness}, in which the prediction network
    receives occupancy grid as an impenetrable space of the object;
  \item {\bf Joint optimization of multi-object poses}, in which the scene configuration
    with multiple objects is evaluated and updated with differentiable collision check;
  \item {\bf Full integration of fusion and 6D pose} as a real-time system, in which the
    object-level volumetric map is exploited for incremental and accurate pose estimation.
\end{itemize}

%% file: related_work/related_work.tex
\section{Related Work}

\begin{figure*}[htbp]
  \centering
  \subfloat[Scene]{
    \includegraphics[width=0.19\linewidth]{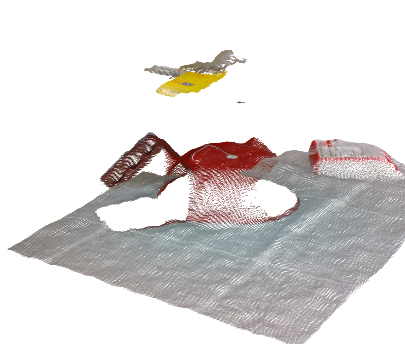}
  }
  \subfloat[Self ($g^{\mathtt{self}}$)]{
    \includegraphics[width=0.19\linewidth]{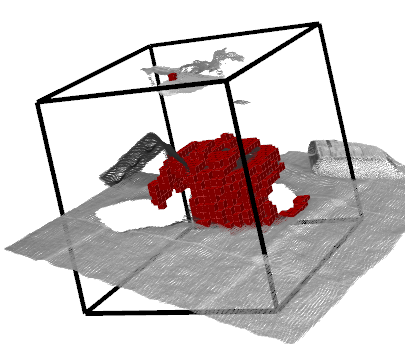}
    \label{figure:surrounding_information:self}
  }
  \subfloat[Other ($g^{\mathtt{other}}$)]{
    \includegraphics[width=0.19\linewidth]{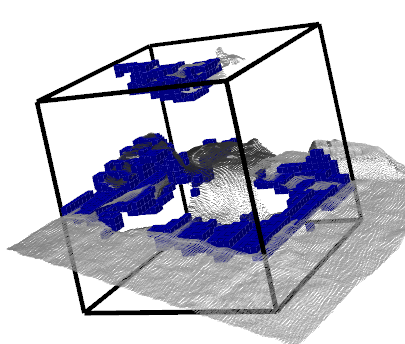}
    \label{figure:surrounding_information:other}
  }
  \subfloat[Free ($g^{\mathtt{free}}$)]{
    \includegraphics[width=0.19\linewidth]{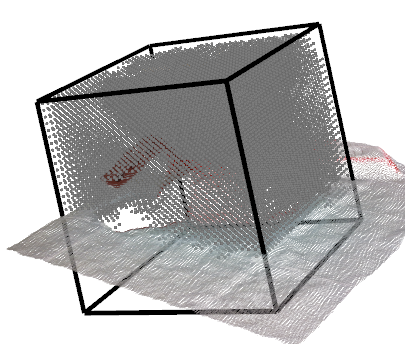}
    \label{figure:surrounding_information:free}
  }
  \subfloat[Unknown ($g^{\mathtt{unknown}}$)]{
    \includegraphics[width=0.19\linewidth]{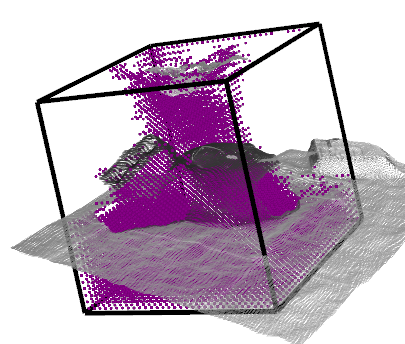}
    \label{figure:surrounding_information:unknown}
  }
  \caption{
    \textbf{Surrounding spatial information.}
    These figures show the occupancy grid ($32\times32\times32$ voxels) of the red bowl.
    The free (d) and unknown (e) grids are visualized with points instead of cubes for visibilities.
  }
  \label{figure:surrounding_information}
  \vspace{2mm} \hrule \vspace{-4mm}
\end{figure*}

\textbf{Template-based} methods \cite{Huttenlocher:etal:PAMI1993, Steger:JPRS2001, Hinterstoisser:etal:CVPR2011, Hinterstoisser:etal:PAMI2012, Hinterstoisser:etal:ACCV2012, Rios-Cabrera:Tuytelaars:ICCV2013} are one of the earliest approaches to pose estimation. Traditionally, these methods involve generating templates by collecting images of the object (or a 3D model) from varying viewpoints in an offline training stage and then scanning the template across an image to find the best match using a distance measure. These methods are sensitive to clutter, occlusions, and lighting conditions, leading to a number of false positives, which in turn requires greater post processing.
\textbf{Sparse feature-based} methods have been a popular alternative to template-based methods for a number of years \cite{Lowe:CVPR2001, Nister:Stewenius:CVPR2006, Philbin:etal:CVPR2007}. These methods are concerned with extracting scale invaraint points of interest from images, describing them with local descriptors, such as SIFT \cite{Lowe:IJCV2004} or SURF \cite{Bay:etal:CVIU2008}, and then storing them in a database to be later matched with at test time to obtain a pose estimate using a method such as RANSAC \cite{Fischler:Bolles:ACM1981}. This processing pipeline can be seen in manipulation tasks, such as MOPED \cite{Collet:etal:IJRR2011}.
With the increase in affordable RGB-D cameras, \textbf{dense} methods have become increasingly popular for object and pose recognition \cite{Drost:etal:CVPR2010, Shotton:etal:CVPR2013, Brachmann:etal:ECCV2014}. These methods involve construction of a 3D point-cloud of a target object, and then matching this with a stored model using popular algorithms such as Iterative Closest Point (ICP) \cite{BeslMckay:PAMI1992}.

The use of deep neural networks is now prevalent in the field of 6D pose
estimation. PoseCNN~\cite{Xiang:etal:RSS2017} was one of the first works to
train an end-to-end system to predict an initial 6D object poses directly from
RGB images, which is then refined using depth-based ICP. Recent RGB-D-based
system are PointFusion~\cite{Xu:etal:CVPR2018} and
DenseFusion~\cite{Wang:etal:CVPR2019}, which individually process the two
sensor modalities (CNNs for RGB, PointNet~\cite{Qi:etal:CVPR2017} for
point-cloud), and then fuse them to extract pixel-wise dense feature
embeddings.
Our work is most closely related to these RGB-D and learning-based approaches
with deep neural networks. In contrast to the point-cloud-based and
target-object-focused approach in the prior work, we process the geometry using
more structured volumetric representation with the geometry information
surrounding the target object.

%% file: methods/methods.tex
\section{MoreFusion}

Our system estimates the 6D pose of a set of known objects given 
RGB-D images of a cluttered scene. We represent 6D poses as a homogeneous
transformation matrix $p \in SE(3)$, and denote a pose as $p = [\matR|\vect]$, where 
$\matR \in SO(3)$ is the rotation and $\vect \in \mathcal{R}^3$ is the translation.

Our system, summarized in \figref{system}, can be divided into four key stages.
(1) An \textbf{object-level volumetric fusion} stage which combines the object instances masks produced from an object detection along with depth measurement and camera tracking component to produce a volumetric map known and unkown objects.
(2) A \textbf{volumetric pose prediction} stage which uses the surrounding information from the volumetric map along with the RGB-D masks to produce an initial pose prediction for each of the objects.
(3) A \textbf{collision-based pose refinement} stage that jointly optimizes the pose of multiple objects via gradient descent by using differentiable collision checking between object CAD models and occupied space from the volumetric map.
(4) A \textbf{CAD alignment} stage that replaces the intermediate representation of each object with a CAD model, containing compact and rich information.
In the following sections, we expend further on each of these stages. 

\begin{figure*}[t]
  \centering
  \includegraphics[width=\linewidth]{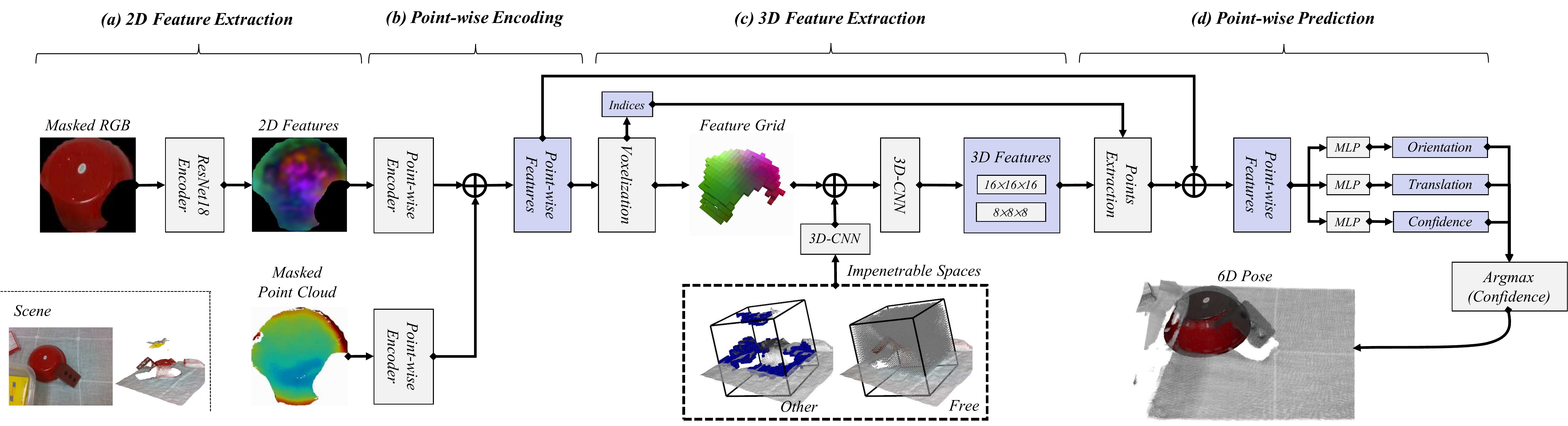}
  \caption{ {\bf Network architecture,} which performs pose prediction using masked RGB-D of the target object
  with its surrounding information as a occupancy grid.}
  \label{figure:network}
\vspace{2mm} \hrule \vspace{-4mm}
\end{figure*}

\input{methods/volumetric_fusion.tex}

\input{methods/pose_prediction.tex}
\input{methods/pose_refinement.tex}

\subsection{CAD Alignment}

After performing the pose estimation and refinement, we spawn object CAD models
into the map once there are enough agreements on the poses estimated in
different views. To compare the object poses estimated in the different camera
coordinate, we first transform those poses into the world coordinate using the
tracked camera pose in camera tracking module (\secref{volumetric_fusion}).
Those transformed object poses are compared using the pose loss, which we also
use for training the pose prediction network (\secref{training_network}). For
the recent $N$ pose hypothesis, we compute the pose loss for each pair, which
gives $N(N-1)$ pose loss: $L_i ~(1 \leq i \leq N(N-1))$. We count how many pose
losses are under the threshold ($L^t$): $M = \mathbf{count}[\![L_i < L^t]\!]$.  When
$M$ reaches a threshold, we initialize the object with that agreed pose.

%% file: methods/volumetric_fusion.tex
\subsection{Object-level Volumetric Fusion} \label{section:volumetric_fusion}

Building a volumetric map is the first stage of our pose estimation system,
which allows the system to gradually increase the knowledge about the scene
until having confidence about object poses in the scene.  For this object-level
volumetric fusion stage, we build a pipeline similar to
~\cite{McCormac:etal:3DV2018,Sunderhauf:etal:IROS2017,Xu:etal:ICRA2019},
combining RGB-D camera tracking, object detection, and volumetric mapping
of detected objects.

\vspace{-1em} \paragraph{RGB-D Camera Tracking} Given that the camera is mounted on the 
end of a robotic arm, we are able to retrieve the accurate pose of the cameras using forward kinematics and a well-calibrated camera. However, to also allow this to be used with a hand-held camera, we adopt the sparse SLAM framework ORB-SLAM2 \cite{Mur-Artal:etal:TRO2017} for camera tracking. Unlike its monocular predecessor \cite{Mur-Artal:Tardos:RSS-MVIGRO2014}, ORB-SLAM2 tracks camera pose in metric space, which is crucial for the volumetric mapping.

\vspace{-1em} \paragraph{Object Detection} Following the prior work
\cite{McCormac:etal:3DV2018,Xu:etal:ICRA2019}, RGB images are passed to
Mask-RCNN\cite{He:etal:ICCV2017} which produce 2D instance masks.

\vspace{-1em} \paragraph{Volumetric Mapping of Detected Objects} We use 
octree-based occupancy mapping, OctoMap\cite{Hornung:etal:AR2013}, 
for the volumetric mapping. By using octree
structure, OctoMap can quickly retrieve the voxel from queried points, which is
critical for both updating the occupancy value from depth measurements and
checking occupancy value when use in the later (pose prediction and refinement)
components of the pipeline.

We build this volumetric map for each detected objects including unknown
(background) objects. In order to track the objects that are already
initialized, we use the intersect over union of the detected mask in the
current frame and rendered mask current reconstruction following prior work
\cite{McCormac:etal:3DV2018,Xu:etal:ICRA2019}. For objects that are already
initialized, we fuse new depth measurements to the volumetric map of that
object, and a new volumetric map is initialized when it finds a new object when
moving the camera. This object-level reconstruction enables to use volumetric representation of
objects as an intermediate representation before the model alignment by pose
estimation.

%% file: methods/pose_prediction.tex
\subsection{Volumetric Pose Prediction} \label{section:pose_prediction}

Our system retrieves surrounding information from
the volumetric map to incorporate spatial awareness of the area surrounding a 
target object into pose prediction. In this section, we describe how this
surrounding information is represented and used in pose prediction.

\subsubsection{Occupancy Grids as Surrounding Information} \label{section:occupancy_grids}

Each target object ($o_\objidx$) for pose prediction carries its own volumetric
occupancy grid. The voxels that make up this grid can be in one of the following states:
(1) Space occupied by the object itself ($g^{\self}$) from the target object reconstruction.
(2) Space occupied by other objects ($g^{\other}$) from reconstruction of the surrounding objects.
(3) Free space ($g^{\mathtt{free}}$) identified by depth measurement.
(4) Unknown space ($g^{\mathtt{unknown}}$) unobserved by mapping because of occlusion and sensor range limit
(\figref{surrounding_information}).

Ideally, the bounding box of surrounding information should cover the whole area of
the target object even if it is occluded. This means the bounding box size
should change depending on the target object size. Since we need to use fixed
voxel dimension for network prediction (e.g., $32\times32\times32$), we use
different voxel size for each object computed from the object model size
(diagonal of the bounding box divided by the voxel dimension).

\subsubsection{Pose Prediction Network Architecture} \label{section:network}

The initial 6D pose of each object is predicted via a deep neural network
(summarized in \figref{network}) that accepts both the occupancy grid
information described in \secref{occupancy_grids} and masked RGB-D images.  The
architecture can be categorized into 4 core components: (1) 2D feature
extraction from RGB from a ResNet; (2) Point-wise encoding of RGB features and
point cloud; (3) Voxelization of the point-wise features followed by 3D-CNNs;
(4) Point-wise pose prediction from both 2D and 3D features.

\vspace{-1em}
\paragraph{2D Feature Extraction from RGB}

Even when depth measurement are available, RGB images can still carry
vital sensor information for precise pose prediction. Because of the
color and texture detail in RGB images, this can be an especially strong
signal for pose prediction of highly-textured and asymmetric objects.

Following \cite{Wang:etal:CVPR2019,Xu:etal:CVPR2018}, we use ResNet18 \cite{He:etal:CVPR2016} 
with succeeding upsampling layers \cite{Zhao:etal:CVPR2017} to extract RGB features from masked images.
Though both prior methods \cite{Wang:etal:CVPR2019,Xu:etal:CVPR2018} used cropped images of objects with a bounding box, we used masked images which makes the network invariant to changes in background
appearance, and also encourages it to focus on retrieving surrounding
information using the occupancy grid.

\vspace{-1em}
\paragraph{Point-wise Encoding of RGB Features and Point Cloud}

Similarly to \cite{Wang:etal:CVPR2019}, both the RGB features and extracted 
point-cloud points (using the target object mask) are encoded via several
fully connected layers to produce point-wise features, which are then concatenated.

\vspace{-1em}
\paragraph{Voxelization and 3D-CNN Processing}

From these point-wise features we build a feature grid (with the
same dimensions as the occupancy grid), which will be combined
with the occupancy grid extracted from the volumetric fusion.  The
concatenated voxel grid is processed by 3D-CNNs to extract hierarchical 3D
features reducing voxel dimension and increasing the channel size.  We process
the original grid (voxel dimension: 32) with 2-strided convolutions to have
hierarchical features (voxel dimension: 16, 8).

An important design choice in this pipeline is to perform 2D feature extraction
before voxelization, instead of directly applying 3D feature extraction on the
voxel grid of raw RGB pixel values.  Though 3D CNNs and 2D CNNs have similar
behaviour when processing RGB-D input, it is hard to use a 3D CNN on a
high resolution grid unlike a 2D image, and also the voxelized grid can have
more missing points than an RGB image because of sensor noise in the depth
image.

\vspace{-1em}
\paragraph{Point-wise Pose Prediction from 2D-3D Features}

To combine the 2D and 3D features for pose prediction, we extract points from
the 3D feature grid that corresponds to the point-wise 2D features with
triliner interpolation.  These 3D and 2D features are concatenated as
point-wise feature vectors for the pose prediction, from which we predict both
the pose and confidence as in \cite{Wang:etal:CVPR2019}.

\vspace{-1em}
\subsubsection{Training the Pose Prediction Network}\label{section:training_network}

\paragraph{Training Loss}

For point-wise pose prediction, we follow DenseFusion \cite{Wang:etal:CVPR2019}
for training loss which is extended version of the model alignment loss from
PoseCNN \cite{Xiang:etal:RSS2017}.  For each pixel-wise prediction, this loss
computes average distance of corresponding points of the object model
transformed with ground truth and predicted pose (pose loss).

Let $[\matR | \vect]$ be ground truth pose, $[\matRhat_i | \vecthat_i]$ be $i$-th
point-wise prediction of the pose, and $\point_\pointidx \in X$ be the point sampled from the object
model.  This pose loss is formulated as:
\begin{eqnarray}
  L_i = \frac{1}{|X|} \sum_\pointidx || (\matR \point_\pointidx + \vect) - (\matRhat_i \point_\pointidx + \vecthat_i) ||.
\end{eqnarray}
For symmetric objects, which have ambiguity for the
correspondence in object model, nearest neighbor of transformed point is used
as the correspondence (symmetric pose loss):
\begin{eqnarray}
  \label{equation:pose_loss}
  L_i = \frac{1}{|X|} \sum_\pointidx \min_{\point_{\pointidx^\prime} \in X} || (\matR \point_\pointidx + \vect) - (\matRhat_i \point_{\pointidx^\prime} + \vecthat_i) ||.
\end{eqnarray}
The confidence of the pose prediction is trained with these pose loss in an unsupervised way.
Let $N$ be number of pixel-wise predictions and $c_i$ be the $i$-th predicted confidence.
The final training loss $L$ is formulated as:
\begin{eqnarray}
  \label{equation:symmetric_pose_loss}
  L = \frac{1}{N} \sum_i (L_i c_i - \lambda \log(c_i)),
\end{eqnarray}
where $\lambda$ is the regularization scaling factor (we use $\lambda=0.015$ following \cite{Wang:etal:CVPR2019}).

\vspace{-1em}
\paragraph{Local Minima in Symmetric Pose Loss}

\begin{figure}[htbp]
  \vspace{-5mm}
  \centering
  \subfloat[Scene]{
    \includegraphics[width=0.31\linewidth]{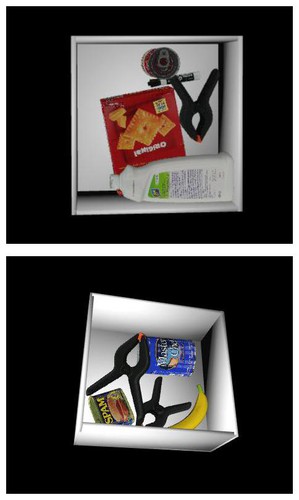}
  }
  \subfloat[Symmetric pose loss]{
    \includegraphics[width=0.31\linewidth]{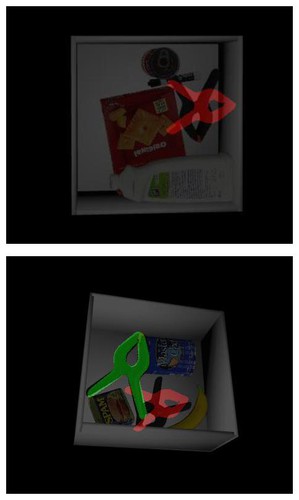}
    \label{figure:symmetric_loss_issue:adds_only}
  }
  \subfloat[With loss warm-up]{
    \includegraphics[width=0.31\linewidth]{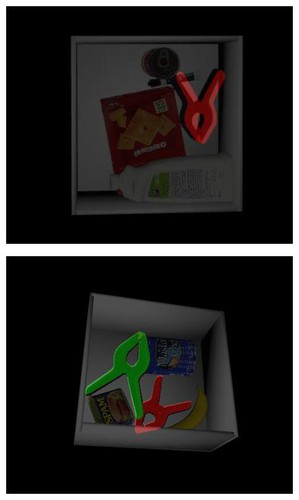}
    \label{figure:symmetric_loss_issue:add_then_adds}
  }
  \caption{{\bf Avoiding local minima with loss warm-up.}
    Our loss warm-up (c) gives much better pose estimation for complex-shaped
    (\eg non-convex) symmetric objects, for which symmetric pose loss (b) is
    prone to local minima.
  }
  \label{figure:symmetric_loss_issue}
  \vspace{2mm} \hrule \vspace{-4mm}
\end{figure}

Though the symmetric pose loss is designed to handle symmetric objects using
nearest neighbour search, we found that this loss is prone to be stuck to local
minima compared to the standard pose loss, which uses 1-to-1 ground truth
correspondence in the object model. \figref{symmetric_loss_issue:adds_only}
shows the examples where the symmetric pose loss has a problem with the local minima
with the non-convex shaped object.

For this issue, we introduce warm-up stage with standard pose loss (\eg 1
epoch) during training before switching to symmetric pose loss.  This training
strategy with warm-up allows the network first to be optimized for the pose
prediction without local minima problem though ignoring symmetries, and then to
be optimized considering the symmetries, which gives much better results for
pose estimation of complex-shaped symmetric objects
(\figref{symmetric_loss_issue:add_then_adds}).

%% file: methods/pose_refinement.tex
\subsection{Collision-based Pose Refinement}

In the previous section, we showed how we combine image-based object
detections, RGB-D data and volumetric estimates of the shapes of
nearby objects to make per-object pose predictions from a network
forward pass. This can often give good initial pose estimates, but not
necessarily a mutually consistent set of estimates for objects which
are in close contact with each other. In this section we therefore
introduce a test-time pose refinement module that can jointly
optimize the poses of multiple objects.

For joint optimization, we introduce differentiable collision checking, composing of
occupancy voxelization of the object CAD model and an intersection
loss between occupancy grids. As both are differentiable, it allows us to optimize
object poses using gradient descent with optimized batch operation on a GPU. 

\vspace{-1em}
\paragraph{Differentiable Occupancy Voxelization}

The average voxelization of feature vectors mentioned in \secref{network}
uses feature vectors using points and is differentiable with respect to
the feature vector. In contrast, the occupancy voxelization needs to be differentiable with
respect to the points. This means the values of each voxel in the occupancy
grid must be a function of the points, which has been transformed by estimated
object pose.

Let $\point_\pointidx$ be a point, $\voxelsize$ be the voxel size, and $\voxelorig$ be the origin
of the voxel (\ie left bottom corner of the voxel grid).
We can transform the point into voxel coordinate with:
\begin{eqnarray}
  \pointtovoxel_\pointidx = (\point_\pointidx - \voxelorig) / \voxelsize.
\end{eqnarray}
For each voxel $\voxel_\voxelidx$ we compute the distance $\delta$ against the point:
\begin{eqnarray}
  \delta_{\pointidx \voxelidx} = ||\pointtovoxel_\pointidx - \voxel_\voxelidx ||.
\end{eqnarray}
We decide the occupancy value based proportional to the distance from nearest point,
resulting in the occupancy value $\occupancy_\voxelidx$ of $\voxelidx$-th voxel being computed as:
\begin{align}
  \delta_\voxelidx &= \min(\delta^t, \min_{\pointidx}(\delta_{\pointidx \voxelidx})) \\
  \occupancy_\voxelidx &= 1 - (\delta_\voxelidx / \delta^t),
\end{align}
where $\delta^t$ is the distance threshold.

\vspace{-1em}
\paragraph{Occupancy Voxelization for a Target Object}

This differentiable occupancy voxelization gives occupancy grids from object
model and hypothesized object pose. For a target object $\obj_\objidx$, the points
sampled from its CAD model $\point_\pointidx$ are transformed with the hypothesized
pose $(\matR_\objidx | \vect_\objidx)$: $\point^\mathtt{T}_{\pointidx} =  \matR_\objidx \vecx_{\pointidx} +
\vect_\objidx$, from which the occupancy value is computed.  The point is uniformly
sampled from the CAD model (including internal part), and gives a hypothesized
occupancy grid of the target object $g^\mathtt{target}_\objidx$.

Similarly, we perform this voxelization with the surrounding objects $\nontargetobj_\nontargetobjidx$.
Unlike the target object voxelization, surrounding objects $\nontargetobj_\nontargetobjidx$ are
voxelized in the voxel coordinate of the target: 
$\pointtovoxel^\nontargetobj_\pointidx = (\point^\nontargetobj_\pointidx - \voxelorig_\obj) / \voxelsize_\obj$
where $\voxelorig_\obj$ is the occupancy grid origin of the target object and $\voxelsize_\obj$ is its voxel size.
This gives the hypothesized occupancy grids of surrounding objects of the
target object: $g^\mathtt{nontarget}_{\nontargetobjidx}$.

\vspace{-1em}
\paragraph{Intersection Loss for Collision Check}

The occupancy voxelization gives the hypothesized occupied space of the target
$g^{\mathtt{target}}_\objidx$ ($\objidx$-th object in the scene) and surrounding objects
$g^{\mathtt{nontarget}}_{\nontargetobjidx}$. The occupancy grids of surrounding objects are
built in the voxel coordinate (center, voxel size) of the target object and
aggregated with element-wise max:
\begin{align}
  g^{\mathtt{nontarget}}_\objidx = \max_\nontargetobjidx g^{\mathtt{nontarget}}_{\nontargetobjidx}.
\end{align}
This gives a single impenetrable occupancy
grid, where the target object pose is penalized with intersection.  In addition
to the impenetrable occupancy grid from the pose hypothesis of surrounding
objects, we also use the occupancy information from the volumetric fusion:
occupied space including background objects $g_\objidx^\other$, free space $g_\objidx^\free$
(\figref{surrounding_information}), as additional impenetrable area:
$g_\objidx^\impen = g_\objidx^\other \cup g_\objidx^\free$.
The collision penalty loss $L_i^{\mathtt{c-}}$ is computed as the intersection between hypothesized
occupied space of the target and the impenetrable surrounding grid:
\begin{align}
  g_\objidx^\mathtt{target-} &= \max_\voxelidx (g^\mathtt{nontarget}_\objidx, g^\impen_\objidx) \\
  L_\objidx^{\mathtt{c+}} &= (g^{\mathtt{target}}_\objidx \odot g^\mathtt{target-}_\objidx)) / \sum_\voxelidx g^{\mathtt{target}}_\objidx,
\end{align}
where $\odot$ is element-wise multiplication.

Though this loss correctly penalizes the collision among the target and
surrounding objects, optimizing for this alone is not enough, as it does
not take into account the visible surface constraint of the target
object $g^\self_\objidx$. The other term in the loss is the intersection between the
hypothesized occupied space of the target and with this grid $L_\objidx^{\mathtt{c+}}$,
to encourage the surface intersection between object pose hypothesis and volumetric
reconstruction:
\begin{eqnarray}
  L_\objidx^{\mathtt{c-}} = (g_\objidx^\mathtt{target} \odot g_\objidx^\self) / \sum_\voxelidx g_\objidx^\self.
\end{eqnarray}
We compute these collision and surface alignment losses for $N$ number of objects
with the batch operation on GPU, and sum them as the total loss $L$:
\begin{eqnarray}
  L = \frac{1}{N} \sum_\objidx (L_\objidx^\mathtt{c+} - L_\objidx^\mathtt{c-}).
\end{eqnarray}
This loss is minimized with gradient
descent allowing us to jointly optimize the pose hypothesis of multiple
objects.

%% file: experiments/experiments.tex
\begin{figure*}[t]
  \centering
  \subfloat[Scene]{
    \includegraphics[width=0.190\linewidth]{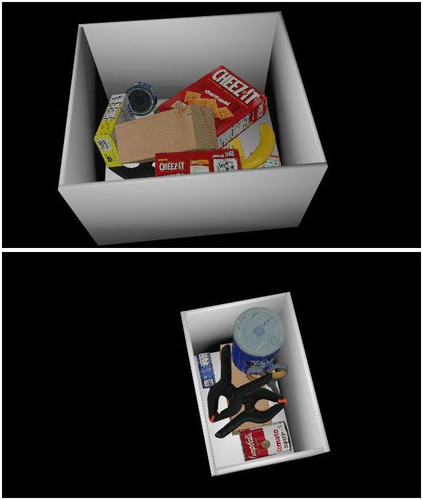}
  } \unskip \vrule height 40mm
  \subfloat[DenseFusion$^*$]{
    \includegraphics[width=0.190\linewidth]{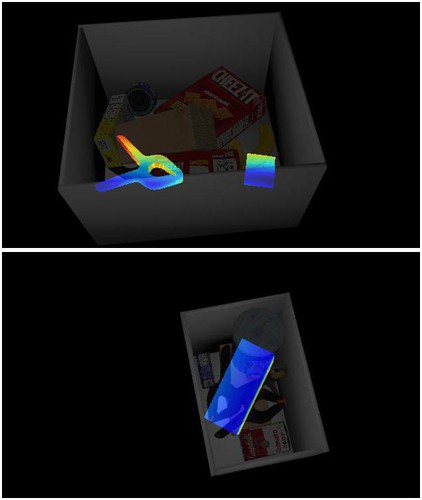}
  } \hspace{-2mm}
  \subfloat[MoreFusion$^\mathtt{-occ}$]{
    \includegraphics[width=0.190\linewidth]{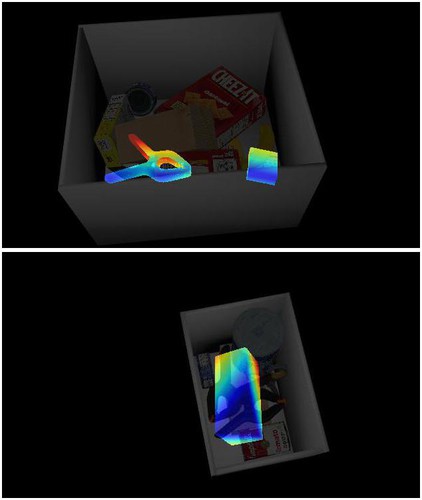}
  } \hspace{-2mm}
  \subfloat[MoreFusion]{
    \includegraphics[width=0.190\linewidth]{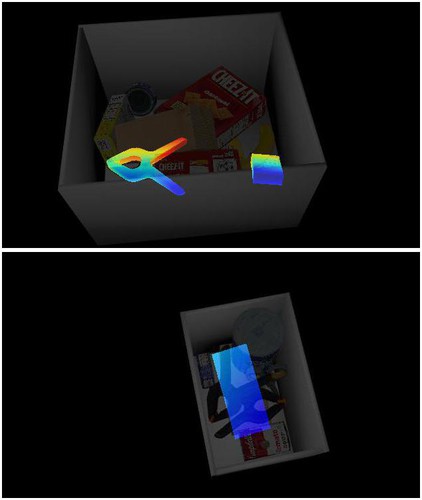}
  } \unskip \vrule height 40mm
  \subfloat[Ground Truth]{
    \includegraphics[width=0.190\linewidth]{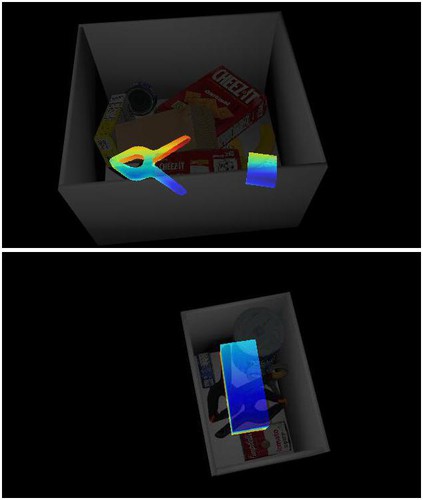}
  }
  \caption{ {\bf Pose prediction with severe occlusions}. Our proposed model
    (MoreFusion) performs consistent pose prediction with surroundings,
    where the baseline (DenseFusion$^*$) and the variant
    without occupancy information (Morefusion$^\mathtt{-occ}$) fails.}
  \label{figure:pose_prediction_results}
  \vspace{2mm}\hrule\vspace{-4mm}
\end{figure*}

\section{Experiments} \label{section:experiments}

In this section, we first evaluate how well the pose prediction
(\secref{pose_prediction_evaluation}) and refinement
(\secref{pose_refinement_evaluation}) performs on 6D pose estimation datasets.
We then demonstrate the system running on a robotic pick-and-place task(\secref{demonstration}).

\subsection{Experimental Settings}

\paragraph{Dataset}
We evaluate our pose estimation components using 21 classes of YCB objects \cite{Calli:etal:ICAR2015}
used in YCB-Video dataset \cite{Xiang:etal:RSS2017}.
YCB-Video dataset has been commonly used for evaluation of 6D pose estimation in prior work,
however, since all of the scenes are table-top,
this dataset is limited in terms of the variety of object orientations and occlusions.

To make the evaluation possible with heavy occlusions and arbitrary
orientations, we built our own synthetic dataset: Cluttered YCB
(\figref{pose_prediction_results}).  We used a physics simulator
\cite{Coumans:etal:2013} to place object models with feasible configurations
from random poses.  This dataset has 1200 scenes ($\mathtt{train:val}=5:1$) and
15 camera frames for each.

\vspace{-1em} \paragraph{Metric} We used the same metric as prior work
\cite{Xiang:etal:RSS2017,Wang:etal:CVPR2019}, which evaluates the average
distance of corresponding points: ADD, ADD-S.  ADD uses ground truth and ADD-S
uses nearest neighbours as correspondence with transforming the model with the
ground truth and estimated pose.  These distances are computed for each object
pose in the dataset, and plotted with the error threshold in x-axis and the
accuracy in the y-axis. The metric is the area under the curve (AUC) using 10cm
as maximum threshold for x-axis.

\subsection{Evaluation of Pose Prediction} \label{section:pose_prediction_evaluation}

\paragraph{Baseline Model}

We used DenseFusion \cite{Wang:etal:CVPR2019} as a baseline model.  For fair
comparison with our proposed model, we reimplemented DenseFusion and trained
with the same settings (\eg data augmentation, normalization, loss).

\tabref{baseline_model} shows the pose prediction result on YCB-Video dataset
using the detection mask of \cite{Xiang:etal:RSS2017}, where
\textit{DenseFusion} is the official GitHub implementation
\footnote{\url{https://github.com/j96w/DenseFusion}} and
\textit{DenseFusion}$^*$ is our version, which includes the warm-up loss
(\secref{training_network}) and the centralization of input point cloud
(analogues to the voxelization step in our model).  We find that the addition
of the two added components leads to big performance improvements. In the
following evaluations, we use DenseFusion$^*$ as the baseline model.

\begin{table}[htbp]
  \small
  \centering
  \caption{{\bf Baseline model results} on YCB-Video dataset, where DenseFusion is the
  official implementation and DenseFusion$^*$ is our reimplemented version.}
  \label{table:baseline_model}
  \begin{tabular}{l|cc}
           Model        &  ADD(-S)          & ADD-S \\ \hhline{=|==}
    DenseFusion                  &  83.9             & 90.9  \\
    DenseFusion$^*$              &  89.1             & 93.3  \\
  \end{tabular}
  \vspace{-4mm}
\end{table}

\paragraph{Results} We compared the proposed model (Morefusion) with the
baseline model (DenseFusion$^*$). For fair comparison, both models predict
object poses in a single-view, where Morefusion only uses occupancy information
from the single-view depth observation. We trained models using combined
dataset of Cluttered-YCB and YCB-Video dataset and tested separately with
ground truth masks.  The result (\tabref{pose_prediction_results},
\figref{pose_prediction_results}) shows that Morefusion consistently predicts
better poses with volumetric CNN and surrounding occupancy information. Larger
improvement is performed on heavily occluded objects (visibility$<$30\%).

\begin{table}[htbp]
  \vspace{-2mm}
  \small
  \centering

  \caption{ {\bf Pose prediction comparison}, where the models are
  trained with the combined dataset and tested separately.}

  \label{table:pose_prediction_results}
  \begin{tabular}{lc|cc}
           Model                & Test Dataset                                                              & ADD(-S)          & ADD-S       \\  \hhline{==|==}
    DenseFusion$^*$             & \multirow{2}{*}{\small YCB-Video}                                         & 88.4             & 94.9        \\
    MoreFusion                  &                                                                           & {\bf 91.0}       & {\bf 95.7}  \\  \hline
    DenseFusion$^{*}$           & \multirow{2}{*}{\small Cluttered YCB}                                     & 81.7             & 91.7        \\
    MoreFusion                  &                                                                           & {\bf 83.4}       & {\bf 92.3}  \\  \hline
    DenseFusion$^*$             & \multirow{2}{*}{\shortstack{\small Cluttered YCB\\(visibility$^{<0.3}$)}} & 59.7             & 83.8        \\
    MoreFusion                  &                                                                           & {\bf 63.5}       & {\bf 85.1}  \\
  \end{tabular}
  \vspace{-2mm}
\end{table}

\begin{table}[t]
  \small
  \centering
  \caption{ {\bf Effect of occupancy information} tested on Cluttered-YCB dataset with the model trained in
  \tabref{pose_prediction_results}.}
  \label{table:pose_prediction_ablation_occupancy}
  \begin{tabular}{l|cc}
    Model                               & ADD(-S) & ADD-S \\  \hhline{=|==}
    DenseFusion$^*$                     & 81.7    & 91.7  \\  \hline
    MoreFusion$^\mathtt{-occ}$          & 82.5    & 91.7  \\
    MoreFusion                          & 83.4    & 92.3  \\
    MoreFusion$^\mathtt{+target^-}$     & 84.7    & 93.3  \\
    MoreFusion$^\mathtt{+target^- +bg}$ & 85.5    & 93.8  \\
  \end{tabular}
  \vspace{-4mm}
\end{table}

To evaluate the effect of surrounding occupancy as input, we tested the trained
model (MoreFusion) feeding different level of occupancy information: discarding
the occupancy information from the single-view observation \textit{-occ}; full
reconstruction of non-target objects \textit{$+target^-$}; full reconstruction
of background objects \textit{+bg}.
\tabref{pose_prediction_ablation_occupancy} shows that the model gives better
prediction as giving more and more occupancy information, which is very common
in our incremental and multi-view object mapping system.  This comparison also
shows that even without occupancy information (Morefusion$^\mathtt{-occ}$) our
model performs better than DenseFusion$^*$ purely because of the 3D-CNNs
architecture.

\begin{figure}[htbp]
  \vspace{-4mm}
  \centering
  \subfloat[No Refinement]{
    \includegraphics[width=0.32\linewidth]{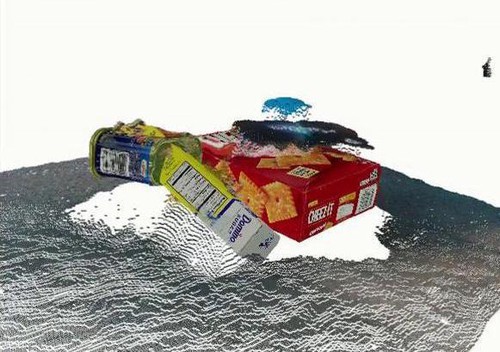}
    \label{figure:pose_refinement_qualitative_init}
  }
  \subfloat[ICP Refinement]{
    \includegraphics[width=0.32\linewidth]{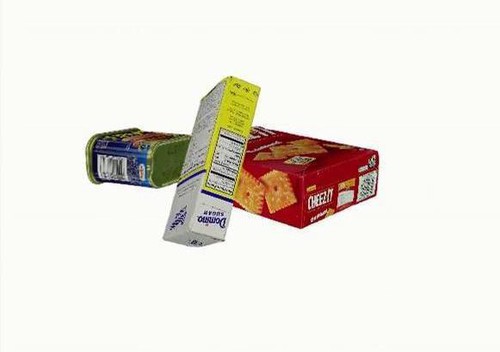}
    \label{figure:pose_refinement_qualitative_icp}
  }
  \subfloat[ICC Refinement]{
    \includegraphics[width=0.32\linewidth]{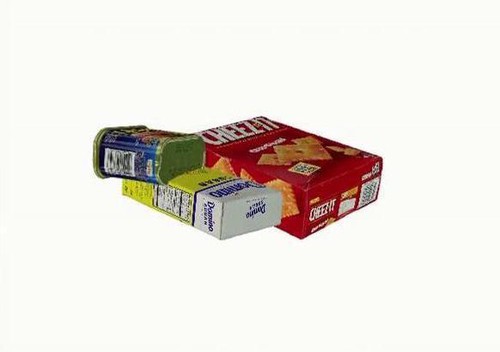}
    \label{figure:pose_refinement_qualitative_icc}
  }
  \caption{{\bf Pose refinement from intersecting object poses},
    where we compare the proposed Iterative Collision Check (ICC)
    against Iterative Closest Point (ICP).
  }
  \label{figure:pose_refinement_qualitative}
  \vspace{2mm} \hrule \vspace{-4mm}
\end{figure}

\begin{figure*}[t]
  \centering
  \includegraphics[width=0.95\linewidth]{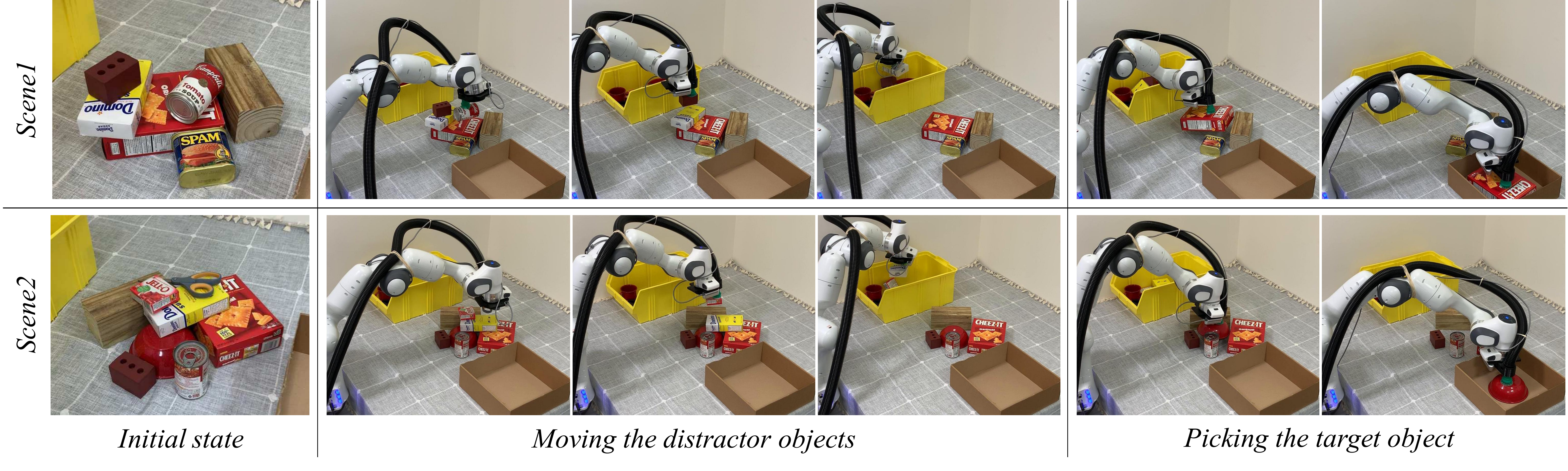}
  \vspace{-2mm}
  \caption{ {\bf Targeted pick-and-place demonstration}, where the robot must move the
  obstructing objects to the container, pick the target object, and then place it 
  in the cardboard box.}
  \label{figure:demonstration}
  \vspace{2mm} \hrule \vspace{-4mm}
\end{figure*}

\subsection{Evaluation of Pose Refinement}  \label{section:pose_refinement_evaluation}

We evaluate our pose refinement, \textit{Iterative Collision Check} (ICC), against
point-to-point Iterative Closest Point (ICP) \cite{BeslMckay:PAMI1992}. Since
ICP only uses masked point cloud of the target object without any reasoning
with surrounding objects, the comparison of ICC with ICP allows us to evaluate
how well and in what case the surrounding-object geometry used in ICC helps
pose refinement in particular.

\figref{pose_refinement_qualitative} shows a typical example where the pose
prediction has object-to-object intersections because of less visibility of the
object (e.g., yellow box). ICC refines object poses to better configurations
than ICP by using the constraints from nearby objects and free-space
reconstructions.

\begin{figure}[htbp]
  \vspace{-2mm}
  \centering
  \includegraphics[width=0.95\linewidth]{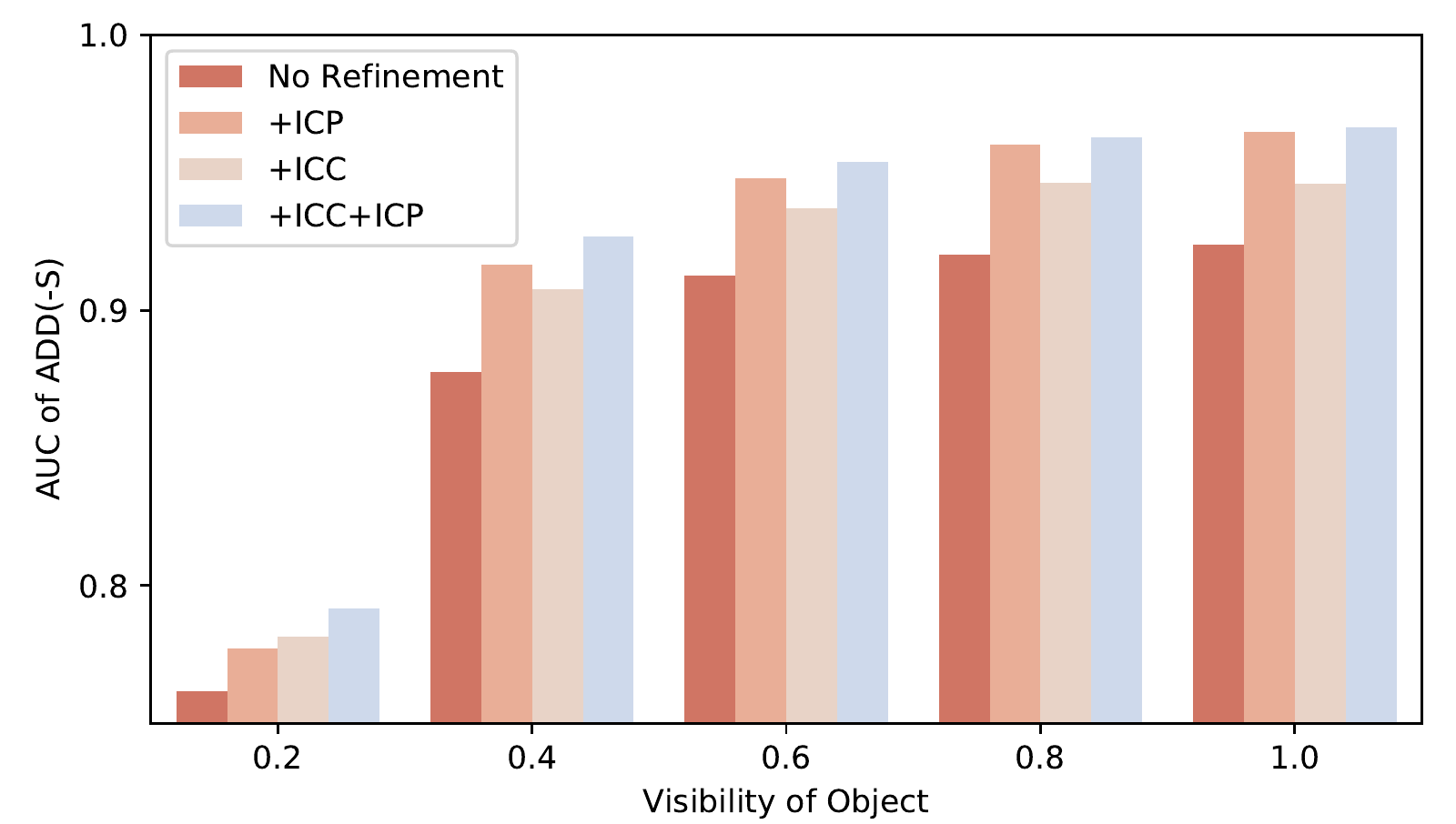}
  \caption{ {\bf Pose refinement results}  on Cluttered YCB-Video, where the
  proposed Iterative Collision Check (ICC) gives best pose estimate combined with ICP.}
  \label{figure:pose_refinement_results}
  \vspace{2mm}\hrule\vspace{-4mm}
\end{figure}

For quantitaive evaluation, we used Cluttered YCB-Video dataset with pose
estimate refined from initial pose prediction MoreFusion in
\tabref{pose_prediction_results}.  \figref{pose_refinement_results} shows how
the metric varies with different visibility on the dataset, in which the
combination of the two methods (+ICC+ICP) gives consistently better pose than
the others. With small occlusions (visibility $>=40\%$), ICC does not perform
as well as ICP because of the discrimination by the voxelization (we use 32
dimensional voxel grid). However, results are at their best with the
combination of the two optimization, where ICC resolves collisions in
discritized space and then ICP aligns surfaces more precisely.

\subsection{Full System Demonstration} \label{section:demonstration}

\begin{figure}[]
  \centering
  \includegraphics[width=0.95\linewidth]{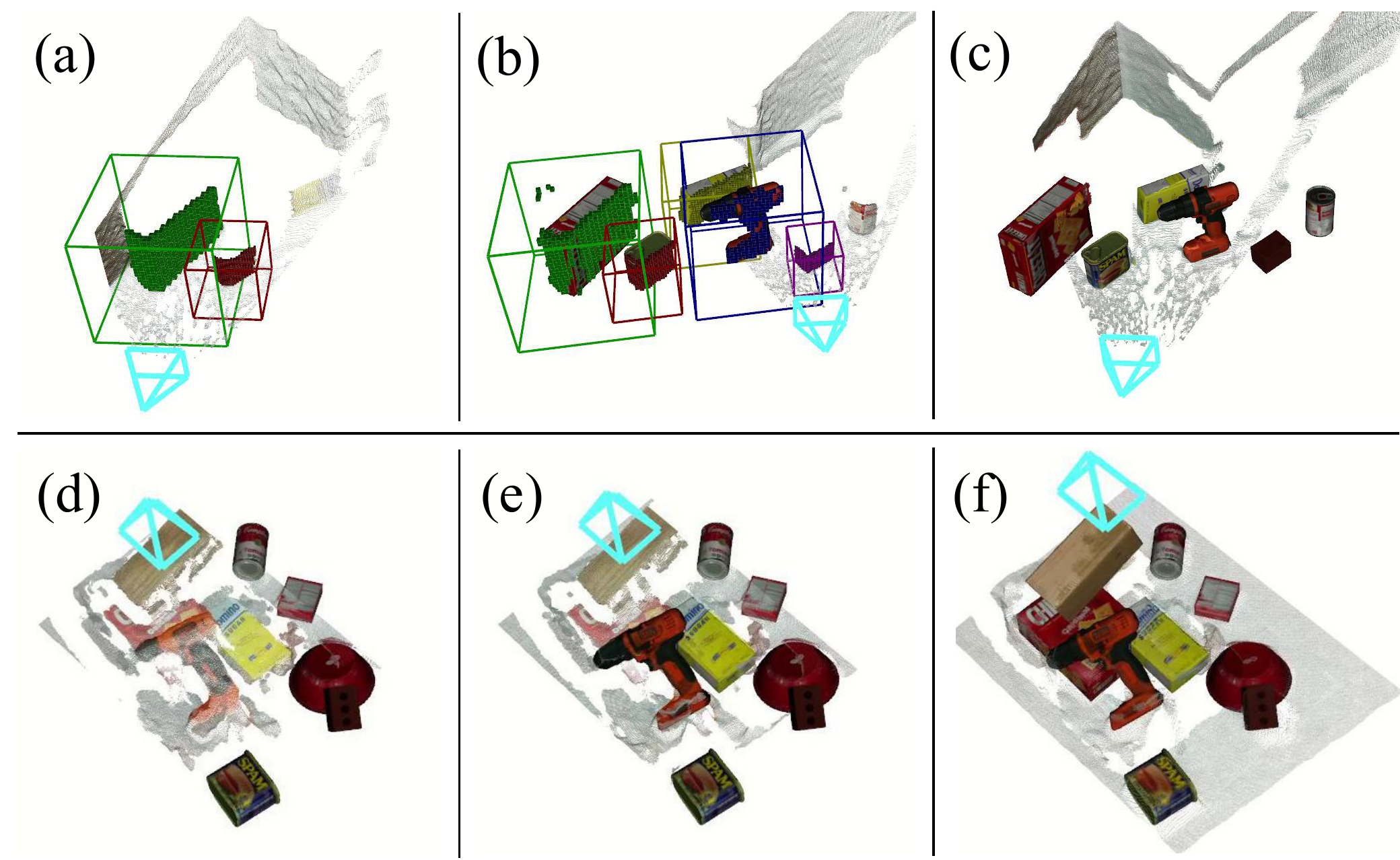}
  \vspace{-2mm}
  \caption{{\bf Real-time full reconstruction.}
    Our system gradually increases the knowledge about the scene with
    volumetric fusion (a) and incremental CAD alignment (b) for the final reconstruction (c).
    The pose hypothesis of surrounding objects (\eg drill, yellow box) are
    utilized to refine the pose predictions, to perform pose estimation
    of heavily occluded objects (\eg red box) (d)-(f).
  }
  \label{figure:reconstruction}
  \vspace{2mm} \hrule \vspace{-4mm}
\end{figure}

We demonstrate the capability of our full system, MoreFusion, with two
demonstration: scene reconstruction, in which the system detects each known
objects in the scene and aligns the pre-build object model (shown in
\figref{reconstruction}); and secondly, a robotic pick-and-place tasks, where
the robot is requested to pick a target object from a cluttered scene with
intelligently removing distractor objects to access the target object (shown in
\figref{demonstration}).

%% file: conclusions/conclusions.tex
\section{Conclusions}

We have shown consistent and accurate pose estimation of objects that may be
heavily occluded by and/or tightly contacting with other objects in cluttered
scenes. Our real-time and incremental pose estimation system
builds an object-level map
that describes the full geometry of objects in the scene, which enables a robot
to manipulate objects in complicated piles with intelligent of dissembling of
occluding objects and oriented placing.
We believe that there is still a long way to go in using known object
models to make persistent models of difficult, cluttered scenes. One
key future direction is to introduce physics reasoning into our
optimization framework.

%% file: acknowledgements/acknowledgements.tex
\section*{Acknowledgments}
Research presented in this paper has been supported by
Dyson Technology Ltd.

%% file: main.bbl
\begin{thebibliography}{10}\itemsep=-1pt

\bibitem{Bay:etal:CVIU2008}
H. Bay, A. Ess, T. Tuytelaars, and L.~Van Gool.
\newblock {{SURF}: Speeded Up Robust Features}.
\newblock {\em {Computer Vision and Image Understanding ({CVIU})}},
  110(3):346--359, 2008.

\bibitem{BeslMckay:PAMI1992}
P. Besl and N. McKay.
\newblock {A method for Registration of {3D} Shapes.}
\newblock {\em {{IEEE} Transactions on Pattern Analysis and Machine
  Intelligence ({PAMI})}}, 14(2):239--256, 1992.

\bibitem{Brachmann:etal:ECCV2014}
E. Brachmann, A. Krull, F. Michel, S. Gumhold, J. Shotton, and C. Rother.
\newblock {Learning 6D Object Pose Estimation using 3D Object Coordinates}.
\newblock In {\em {Proceedings of the European Conference on Computer Vision
  ({ECCV})}}, 2014.

\bibitem{Calli:etal:ICAR2015}
B. Calli, A. Singh, A. Walsman, P Srinivasa S.~and, Abbeel, and A.~M. Dollar.
\newblock The ycb object and model set: Towards common benchmarks for
  manipulation research.
\newblock In {\em International Conference on Advanced Robotics (ICAR)}, pages
  510--517, 2015.

\bibitem{Collet:etal:IJRR2011}
Alvaro Collet, Manuel Martinez, and Siddhartha~S Srinivasa.
\newblock The moped framework: Object recognition and pose estimation for
  manipulation.
\newblock {\em {International Journal of Robotics Research ({IJRR})}},
  30(10):1284 -- 1306, 2011.

\bibitem{Coumans:etal:2013}
Erwin Coumans et~al.
\newblock Bullet physics library.
\newblock {\em Open source: bulletphysics. org}, 2013.

\bibitem{Drost:etal:CVPR2010}
B. Drost, M. Ulrich, N. Navab, and S. Ilic.
\newblock {Model globally, match locally: Efficient and robust {3D} object
  recognition}.
\newblock In {\em {Proceedings of the {IEEE} Conference on Computer Vision and
  Pattern Recognition ({CVPR})}}, 2010.

\bibitem{Fischler:Bolles:ACM1981}
M.~A. Fischler and R.~C. Bolles.
\newblock {Random sample consensus: a paradigm for model fitting with
  applications to image analysis and automated cartography}.
\newblock {\em Communications of the ACM}, 24(6):381--395, 1981.

\bibitem{He:etal:ICCV2017}
K. He, G. Gkioxari, P. Doll{\'a}r, and R. Girshick.
\newblock Mask r-cnn.
\newblock In {\em {Proceedings of the International Conference on Computer
  Vision ({ICCV})}}, 2017.

\bibitem{He:etal:CVPR2016}
K. He, X. Zhang, S. Ren, and J. Sun.
\newblock Deep residual learning for image recognition.
\newblock In {\em {Proceedings of the {IEEE} Conference on Computer Vision and
  Pattern Recognition ({CVPR})}}, 2016.

\bibitem{Hinterstoisser:etal:PAMI2012}
S. Hinterstoisser, C. Cagniart, S. Ilic, P. Sturm, N. Navab, P. Fua, and V.
  Lepetit.
\newblock {Gradient Response Maps for Real-Time Detection of Texture-Less
  Objects}.
\newblock {\em {{IEEE} Transactions on Pattern Analysis and Machine
  Intelligence ({PAMI})}}, 2012.

\bibitem{Hinterstoisser:etal:CVPR2011}
S. Hinterstoisser, S. Holzer, C. Cagniart, S. Ilic, K. Konolidge, N. Navab, and
  V. Lepetit.
\newblock {Multimodal Templates for Real-Time Detection of Texture-less Objects
  in Heavily Cluttered Scenes}.
\newblock In {\em {Proceedings of the International Conference on Computer
  Vision ({ICCV})}}, 2011.

\bibitem{Hinterstoisser:etal:ACCV2012}
S. Hinterstoisser, V. Lepetit, S. Ilic, S. Holzer, G. Bradski, K. Konolige, and
  N. Navab.
\newblock {Model-Based Training, Detection and Pose Estimation of Texture-less
  3D Objects in Heavily Cluttered Scenes}.
\newblock In {\em {Proceedings of the Asian Conference on Computer Vision
  ({ACCV})}}, 2012.

\bibitem{Hornung:etal:AR2013}
Armin Hornung, Kai~M. Wurm, Maren Bennewitz, Cyrill Stachniss, and Wolfram
  Burgard.
\newblock {OctoMap}: An efficient probabilistic {3D} mapping framework based on
  octrees.
\newblock {\em {Autonomous Robots}}, 2013.

\bibitem{Huttenlocher:etal:PAMI1993}
Daniel~P. Huttenlocher, Gregory~A. Klanderman, and William~J Rucklidge.
\newblock Comparing images using the hausdorff distance.
\newblock {\em {{IEEE} Transactions on Pattern Analysis and Machine
  Intelligence ({PAMI})}}, 15(9):850--863, 1993.

\bibitem{Lowe:CVPR2001}
D. Lowe.
\newblock {Local Feature View Clustering for 3D Object Recognition}.
\newblock In {\em {Proceedings of the {IEEE} Conference on Computer Vision and
  Pattern Recognition ({CVPR})}}, 2001.

\bibitem{Lowe:IJCV2004}
D.~G. Lowe.
\newblock {Distinctive image features from scale-invariant keypoints}.
\newblock {\em {International Journal of Computer Vision ({IJCV})}},
  60(2):91--110, 2004.

\bibitem{McCormac:etal:3DV2018}
J. McCormac, R. Clark, M. Bloesch, A.~J. Davison, and S. Leutenegger.
\newblock {Fusion\texttt{++}}:volumetric object-level slam.
\newblock In {\em {Proceedings of the International Conference on 3D Vision
  ({3DV})}}, 2018.

\bibitem{Mur-Artal:Tardos:RSS-MVIGRO2014}
R. Mur-Artal and J.~D Tard{\'o}s.
\newblock {ORB-SLAM: Tracking and Mapping Recognizable Features}.
\newblock In {\em {Workshop on Multi View Geometry in Robotics (MVIGRO) - RSS
  2014}}, 2014.

\bibitem{Mur-Artal:etal:TRO2017}
R. Mur-Artal and J.~D. Tard{\'o}s.
\newblock {ORB-SLAM2: An Open-Source SLAM System for Monocular, Stereo, and
  RGB-D Cameras}.
\newblock {\em {{IEEE} Transactions on Robotics ({T-RO})}}, 33(5):1255--1262,
  2017.

\bibitem{Nister:Stewenius:CVPR2006}
D. Nister and H. Stewenius.
\newblock {Scalable Recognition with a Vocabulary Tree}.
\newblock In {\em {Proceedings of the {IEEE} Conference on Computer Vision and
  Pattern Recognition ({CVPR})}}, 2006.

\bibitem{Philbin:etal:CVPR2007}
J. Philbin, O. Chum, M. Isard, J. Sivic, and A. Zisserman.
\newblock {Object Retrieval with Large Vocabularies and Fast Spatial Matching}.
\newblock In {\em {Proceedings of the {IEEE} Conference on Computer Vision and
  Pattern Recognition ({CVPR})}}, 2007.

\bibitem{Qi:etal:CVPR2017}
Charles~R Qi, Hao Su, Kaichun Mo, and Leonidas~J Guibas.
\newblock Pointnet: {D}eep {L}earning on {P}oint {S}ets for 3{D}
  {C}lassification and {S}egmentation.
\newblock In {\em {Proceedings of the {IEEE} Conference on Computer Vision and
  Pattern Recognition ({CVPR})}}, pages 652--660, 2017.

\bibitem{Rios-Cabrera:Tuytelaars:ICCV2013}
Reyes Rios-Cabrera and Tinne Tuytelaars.
\newblock Discriminatively trained templates for 3d object detection: A real
  time scalable approach.
\newblock In {\em {Proceedings of the International Conference on Computer
  Vision ({ICCV})}}, 2013.

\bibitem{Shotton:etal:CVPR2013}
J. Shotton, B. Glocker, C. Zach, S. Izadi, A. Criminisi, and A. Fitzgibbon.
\newblock {Scene coordinate regression forests for camera relocalization in
  RGB-D images}.
\newblock In {\em {Proceedings of the {IEEE} Conference on Computer Vision and
  Pattern Recognition ({CVPR})}}, 2013.

\bibitem{Steger:JPRS2001}
Carsten Steger.
\newblock Similarity measures for occlusion, clutter, and illumination
  invariant object recognition.
\newblock In {\em Joint Pattern Recognition Symposium}, 2001.

\bibitem{Sunderhauf:etal:IROS2017}
N. S\"{u}nderhauf, T.~T. Pham, Y. Latif, M. Milford, and I. Reid.
\newblock Meaningful maps with object-oriented semantic mapping.
\newblock In {\em {Proceedings of the {IEEE/RSJ} Conference on Intelligent
  Robots and Systems ({IROS})}}, 2017.

\bibitem{Wang:etal:CVPR2019}
Chen Wang, Danfei Xu, Yuke Zhu, Roberto Mart{\'\i}n-Mart{\'\i}n, Cewu Lu, Li
  Fei-Fei, and Silvio Savarese.
\newblock {DenseFusion}: {6D} object pose estimation by iterative dense fusion.
\newblock {\em {Proceedings of the {IEEE} Conference on Computer Vision and
  Pattern Recognition ({CVPR})}}, 2019.

\bibitem{Xiang:etal:RSS2017}
Yu Xiang, Tanner Schmidt, Venkatraman Narayanan, and Dieter Fox.
\newblock {PoseCNN}: A convolutional neural network for {6D} object pose
  estimation in cluttered scenes.
\newblock In {\em {Proceedings of Robotics: Science and Systems ({RSS})}},
  2018.

\bibitem{Xu:etal:ICRA2019}
Binbin Xu, Wenbin Li, Dimos Tzoumanikas, Michael Bloesch, Andrew Davison, and
  Stefan Leutenegger.
\newblock Mid-fusion: Octree-based object-level multi-instance dynamic slam.
\newblock In {\em {Proceedings of the {IEEE} International Conference on
  Robotics and Automation ({ICRA})}}, 2019.

\bibitem{Xu:etal:CVPR2018}
Danfei Xu, Dragomir Anguelov, and Ashesh Jain.
\newblock {PointFusion}: Deep sensor fusion for {3D} bounding box estimation.
\newblock In {\em {Proceedings of the {IEEE} Conference on Computer Vision and
  Pattern Recognition ({CVPR})}}, 2018.

\bibitem{Zhao:etal:CVPR2017}
Hengshuang Zhao, Jianping Shi, Xiaojuan Qi, Xiaogang Wang, and Jiaya Jia.
\newblock Pyramid scene parsing network.
\newblock In {\em {Proceedings of the {IEEE} Conference on Computer Vision and
  Pattern Recognition ({CVPR})}}, 2017.

\end{thebibliography}
